\documentclass[conference]{IEEEtran}

\usepackage{times}
\usepackage{epsfig}
\usepackage{graphicx}
\usepackage{amsmath}
\usepackage{amssymb}
\usepackage{multirow}
\usepackage{setspace}
\usepackage{subfigure}
\usepackage{balance}
\usepackage{url}

\usepackage{array}
\newcolumntype{V}{>{$\vcenter\bgroup\hbox\bgroup}c<{\egroup\egroup$}}
\def\Hline{\noalign{\hrule height 4\arrayrulewidth}}

\graphicspath{{./imgs/}}

\begin{document}

\title{Descriptor Transition Tables for Object Retrieval using Unconstrained Cluttered Video Acquired using a Consumer Level Handheld Mobile Device}

\author{
\begin{tabular}{cc}
Warren Rieutort-Louis   & Ognjen Arandjelovi\'c\\
University of Cambridge & University of St Andrews\\
Cambridge               & St Andrews\\
CB2 1TQ                 & KY16 9SX\\
United Kingdom          & United Kingdom\\
                        & \texttt{ognjen.arandjelovic@gmail.com}\\
                        &\\&\\
\end{tabular}}

\maketitle

\begin{abstract}
Visual recognition and vision based retrieval of objects from large databases are tasks with a wide spectrum of potential applications. In this paper we propose a novel recognition method from video sequences suitable for retrieval from databases acquired in highly unconstrained conditions e.g.\ using a mobile consumer-level device such as a phone. On the lowest level, we represent each sequence as a 3D mesh of densely packed local appearance descriptors. While image plane geometry is captured implicitly by a large overlap of neighbouring regions from which the descriptors are extracted, 3D information is extracted by means of a descriptor transition table, learnt from a single sequence for each known gallery object. These allow us to connect local descriptors along the 3$^{\text{rd}}$ dimension (which corresponds to viewpoint changes), thus resulting in a set of variable length Markov chains for each video. The matching of two sets of such chains is formulated as a statistical hypothesis test, whereby a subset of each is chosen to maximize the likelihood that the corresponding video sequences show the same object. The effectiveness of the proposed algorithm is empirically evaluated on the Amsterdam Library of Object Images and a new highly challenging video data set acquired using a mobile phone. On both data sets our method is shown to be successful in recognition in the presence of background clutter and large viewpoint changes.
\end{abstract}

\IEEEpeerreviewmaketitle

\section{Introduction}
Owing to its pervasive application potential, computer based object recognition has been a focus of much computer vision research in the last decade. Successful proof-of-concept as well as commercial applications have been demonstrated in the context of large-scale image retrieval~\cite{AranZiss2011}, urban scene recognition~\cite{ZhouLapeXiaoTorr+2014}, augmented reality, and others. While most existing methods address the problem of object recognition using individual images, in this paper we focus on recognition from video. In other words a sequence of frames (images) of an unknown, query object is matched against a database of sequences of known, gallery objects. This problem setting is of an increasing significance considering the ease with which users can acquire and store videos (e.g.\ using a mobile phone camera and cloud storage), and the recognition robustness that the availability of additional data (in comparison with a single image) can provide (e.g.\ with respect to viewpoint).

\begin{figure}
  \centering
  \subfigure[Sequence 1]{\fbox{\includegraphics[width=0.22\textwidth]{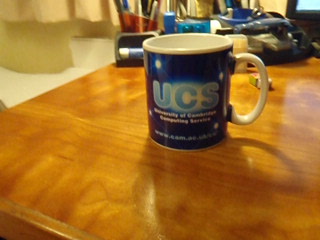}}}~~~~
  \subfigure[Sequence 2]{\fbox{\includegraphics[width=0.22\textwidth]{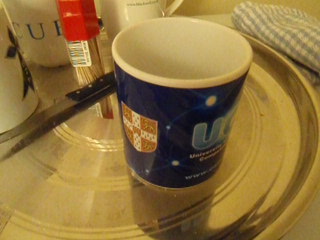}}}
  \subfigure[Sequence 3]{\fbox{\includegraphics[width=0.22\textwidth]{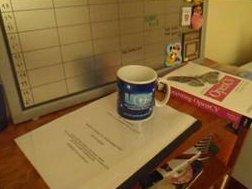}}}~~~~
  \subfigure[Sequence 4]{\fbox{\includegraphics[width=0.22\textwidth]{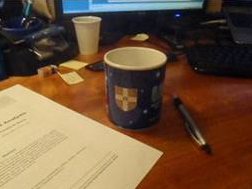}}}
  \caption{Typical frames from videos of the same object, acquired at different times and in different backgrounds. Recognition of untextured (``smooth'') objects across pose and illumination changes, and the presence of clutter poses a major challenge to existing methods.}
  \label{f:challenges}
\end{figure}

\section{Previous work}\label{s:prev}
Automatic object recognition has attracted considerable research effort. Here we briefly review some of the directions taken by previously proposed approaches in the literature.

\subsection{Holistic representations} An important source of difficulties that arise in an attempt to understand the content encoded by pixel intensities is the distributed nature of the information that can be potentially relevant to grouping decisions. The observation that the ultimate interpretation of an image fragment in the context of object recognition more often than not depends on its context, if not on the entire image, motivates the use of holistic representations.

\paragraph{Prototypes}
One of the most common ways of representing objects is by a set of appearance prototypes or exemplars~\cite{KlarJain2013,SiZhu2013}. This representation is simple, directly measurable and can thus be used irrespective of object scale or data quality in general. It also has a clear probabilistic interpretation, which means that any of a number of well understood off-the-shelf statistical methods can be applied to it. The entirety of an object's appearance is effectively described by the underlying probability density function which describes the object's possible appearance variation~\cite{Aran2014,Aran2014b}.

At its core, the representation is in fact the image itself and is thus not invariant to virtually anything at all. Background clutter poses a significant problem, just as does occlusion, as well as in the case of general, non-planar objects, changing viewpoint and illumination. Depending on the nature of the object (planarity and reflectance properties) a large number of exemplars may be needed to capture the entire corpus of appearance variation~\cite{SiZhu2013}.

\paragraph{Model based}
Model based object representation is rather different in nature from the previously discussed prototypes and the only truly view-invariant representation. Rather than describing an object in terms of how it appears in images, an object is characterized by its inherent properties such as shape and texture. As a result, this
representation is not directly measurable from images. Instead, given models of objects of interest, recognition is performed by finding the model that best fits the image using back-projection: using a postulated set of viewing parameters (e.g.\ camera angle and illumination direction) the model is used to predict what the image
should look like, which is then compared to the actual, observed appearance~\cite{FaneDantGallFoss+2013,HejrRama2014,HintLepeIlicHolz+2012}. More generally, it need not be appearances that are compared but rather any measurable image features~\cite{SaueCootTayl2011}. However, due to the constrained nature of this representation, it is generally suitable only for recognition within a narrow class of objects.

\subsection{Local representations} In contrast to holistic approaches, local methods focus on describing different parts of objects first, building the representation of an entire object from bottom up.

\paragraph{Part based representations}
The pictorial structures approach~\cite{EndrShihJiaaHoie2013,YangRama2011} is a typical example from the group of part based representations \cite{PandLaze2011}. Simultaneously using appearance and spatial information, an object is represented by a geometrically deformable configuration of different predefined (and typically manually chosen) parts. Successful examples from the literature include faces (with the eyes, the mouth and the nose as parts), the human body (with the limbs and the head as parts), motorcycles and aeroplanes. In the context of this paper, part based representations suffer from similar limitations as model based approaches.

\paragraph{Local feature based methods}
In general object recognition tasks, part based approaches are largely overshadowed by the success of representations which use local descriptors~\cite{AranZiss2011,Aran2012f,GirsDonaDarrMali2014,RublRabaKonoBrad2011}. The idea is simple: at the lowest level small image patches are represented by feature vectors, which are at higher levels integrated into a consistent object description. Thus, there are three main design areas which have given rise to a variety of methods:
\begin{itemize}
  \item which image patches are considered~\cite{SingGuptEfro2012},
  \item how each patch is represented~\cite{GirsDonaDarrMali2014,RublRabaKonoBrad2011}, and
  \item how local descriptors are used to describe the entirety of an object~\cite{HeikPietSchm2009,JegoDouzSchmPere2010}.
\end{itemize}
In spirit, the method proposed in this paper is local feature based with most of our contributions falling within the scope of the last of the aforementioned design issues.

\section{Proposed method}
Our general approach in recognizing the object in a novel, query image sequence is to compare it to training sequences of all ``known'' objects in the gallery and assign it to the one with the highest degree of similarity. Since the object of interest has unknown, arbitrary shape and appearance, and may be embedded in significant background clutter, extracting a model of the object's appearance from each sequence in isolation for the purpose of comparing model parameters is difficult without imposing constraints on the class, shape, or appearance of the object or the background (as was done for example by Arandjelovi\'c and Zisserman \cite{AranZiss2011} who constrained their attention to sculptures which allowed them to learn and perform super-pixel-level background/foreground segmentation).

Hence in order to avoid the need for overly restrictive assumptions, we take a different approach. We merely assume that the object of interest is roughly in the centre of the video. Then when two sequences are compared with each other (one from the gallery of known objects, the other a query sequence) we seek to find the model parameters which best explain both sequences i.e.\ that automatically infer the common appearance elements between them. Thus, each comparison, even of sequences which correspond to different objects, produces a hypothesised model of an object. The aim is that the hypothesised model produced when correctly matching sequences are compared is that with the highest likelihood. We now explain how each of the components of our algorithm fits into the overall framework which accomplishes this. In summary, our algorithm comprises the following sequence of steps:
\begin{itemize}
  \item Motion parallax based frame-wise scale normalization,
  \item Extraction of low-level spatio-termporal appearance features,
  \item Model parameter fitting via cross-sequence mutual likelihood maximization, and
  \item Quasi-volumetric foreground/background video sequence segmentation and model likelihood estimation.
\end{itemize}

\subsection{Baseline appearance representation}
At the bottom-most level, our method is based on describing small image patches i.e.\ local appearance \cite{BiadAran2015}. This is motivated by observing that if such patches are chosen wisely, they correspond to object parts with consistent geometry and texture, and are thus less sensitive in appearance to variation in viewpoint. For such regions, representations such as the Scale Invariant Features Transform (SIFT)~\cite{Lowe2004} and the related Histograms of Oriented Gradients (HOGs)~\cite{DalaTrig2005} descriptors have been proposed and demonstrated effective in a variety of applications~\cite{Aran2012d}. Being based on image intensity gradients they also show low sensitivity to illumination changes~\cite{Aran2012e,Aran2013,MikoSchm2004}.

Most local descriptor based methods employ descriptors in a sparse fashion by focusing on a set of detected interest points~\cite{RublRabaKonoBrad2011}. When the number of detections is large this can achieve impressive robustness to partial occlusion and image clutter. However, a serious limitation of this approach is that it cannot handle untextured objects~\cite{AranZiss2011,Aran2012i}. A related problem is that of enforcing geometric constraints between local descriptors. If no geometric constraints are used (e.g.\ as in the bag-of-words approach \cite{AranZiss2011,SiviRussEfro+2005}), the representation lacks discriminative power to distinguish between similar objects, especially two objects of the same category, or complex objects with the same basic building element~\cite{Aran2010}. For example, with this representation, both a bicycle and a metal rail fence may end up looking very similar indeed. On the other hand, devising geometric constraints suitable for general, 3D object is challenging. Lowe's use of the Hough
transform effectively restricts the class of objects to nearly-planar ones or, alternatively, restricts camera viewpoint to only very small deviations.

\subsubsection{Capturing view geometry through redundancy}\label{sss:geoFrame}
\begin{figure}[htb]
  \centering
  \subfigure[Dense grid]{\includegraphics[width=0.22\textwidth]{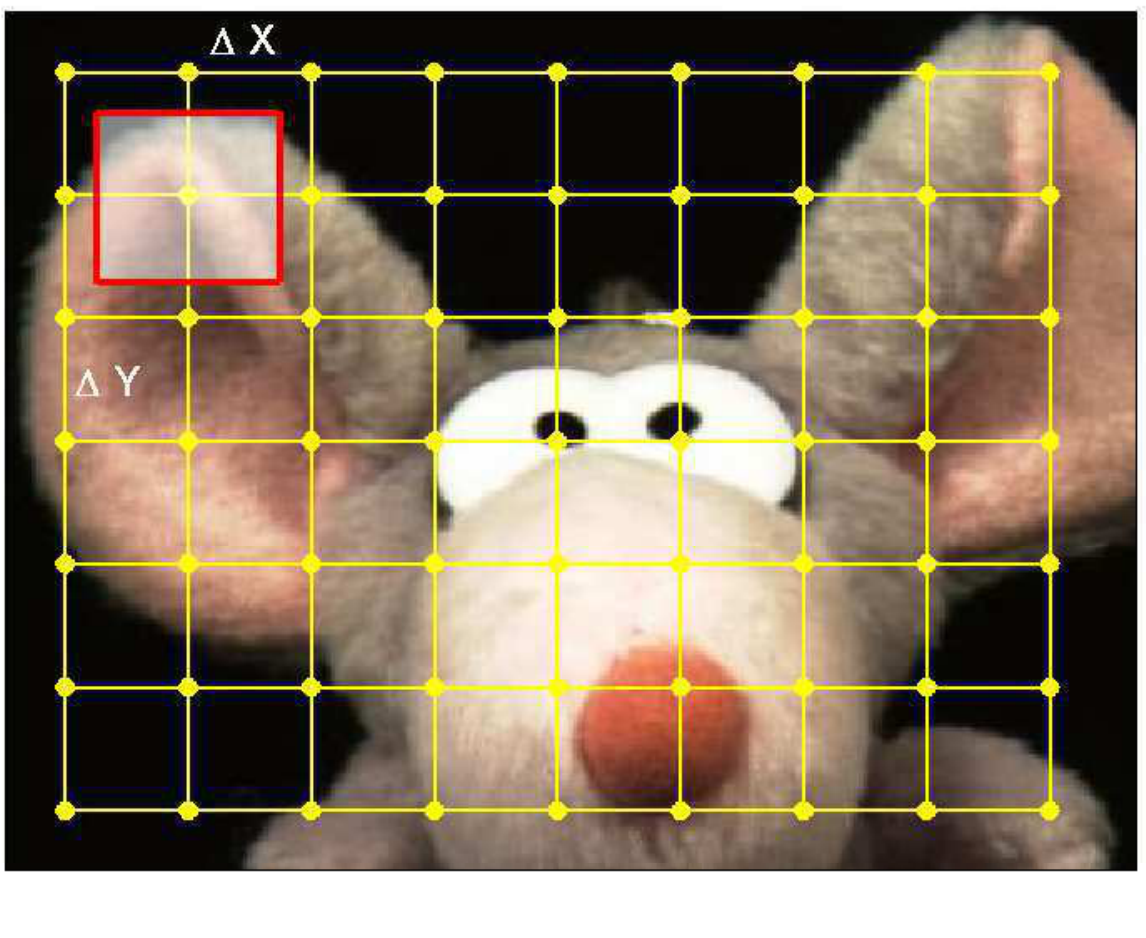}}~~~~~~
  \subfigure[Neighbouring patch overlap]{\includegraphics[width=0.22\textwidth]{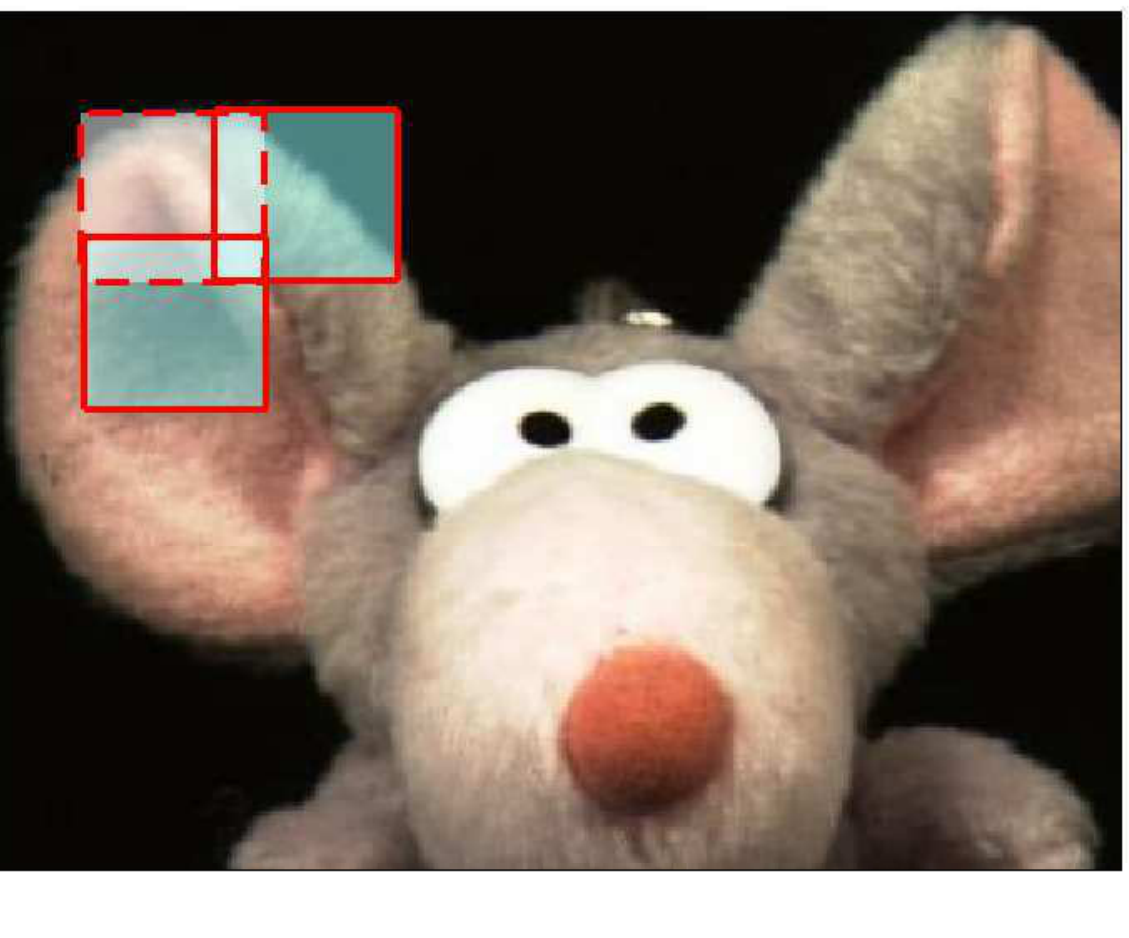}}
  \caption{ (a) We describe the appearance of an entire video sequence, and thus both of the object of interest as well as any present clutter, by collecting local image patches collected over a dense grid. (b) Geometric relationship between patches is captured implicitly by making grid spacing smaller than the patch size. The resulting patch overlap means that the same image region contributes to multiple local descriptors. }
  \label{f:grid}
\end{figure}

We tackle the issue of geometry, that is, geometric constraints between different appearance features, using two complementary approaches. The first of these deals with image plane geometry i.e.\ the relationship between extracted local patches in a single frame. This is realized implicitly, by making the relative shift $(\Delta x,\Delta y)$ between neighbouring patches smaller than their dimensions $s_x$ and $s_y$, i.e.\ $s_x > \Delta x$ and $s_y > \Delta y$, resulting in patch overlap. For our experiments we used 90\% overlap between neighbouring pathes both in the vertical and the horizontal direction:
\begin{align}
\Delta x / s_x = \Delta y / s_y = 0.1,
\end{align}
as illustrated conceptually in Fig~\ref{f:grid}.

To see how this approach captures geometric constraints in the image plane, notice that for any two arbitrary patches there is a sequence of patches, each neighbouring the previous one, that connects them. Since neighbouring patches greatly overlap and objects tend to be smooth, the difference in their appearance is small and the aforementioned patch sequence describes a manifold-like structure in the image space (for a similar idea in a different domain, that of temporal topic modelling, see the methods and analyses in \cite{BeykPhunAranVenk2015,BeykPhunAranVenk2016}). Extending this to the entire set of object image patches collected over our dense grid, it can be seen that this set then describes a 2D surface in the image space. At the same time, the proposed overlap solves the problem of object-grid alignment too. Because our patches are densely packed, while translating the object relative to the grid may change the appearance of any single patch, it leaves the entire set collected over the image unchanged.

\paragraph{Representing local appearance}
Following previous work, we use the SIFT descriptor to describe each patch in our dense grid and then quantize it by assigning it to the nearest of the $k$ clusters, or descriptor words, estimated by $k$-means clustering all descriptors extracted from all frames of the training image sequences (we used $k=500$).

\subsection{Scale normalization}
As already mentioned in Sec~\ref{s:prev} most of the existing local appearance based recognition algorithms are sparse in the sense that they focus on a relatively small number of salient, stable loci. These can be detected using one of a number of keypoint detectors~\cite{Aran2012f,Lind2013}. Considering that all modern keypoint detection algorithms explicitly consider the scale of the keypoint, local descriptors are extracted at the corresponding scale thereby achieving scale invariance. Given that in the proposed method local descriptors are collected over a dense grid, the benefit of scale invariance does not come so readily and requires a preceding normalization stage. Our approach is broadly motion parallax based.

We start by computing the optic flow field, using a variant of the well-known Lucas-Kanade algorithm. This field is modelled as comprising a translatory component (recall that our aim is to handle videos acquired in unconstrained conditions using handheld devices) and a rotational component. To correct for the former, we subtract the mean flow vector computed over a frame from the entire field. Since the remaining flow field is generated by a rotational movement of the camera with the object of interest in the centre of the view, motion parallax effected by the depth differential between the object and the background is demonstrated by a discontinuity in the magnitude of the optic flow field at the object edges. By detecting the rough object boundary based on this discontinuity the rough object size within the frame can be estimated and normalized by re-scaling the frame.

\subsection{Descriptor transition tables}
In Sec~\ref{sss:geoFrame} we explained how the proposed method implicitly captures image plane geometry, that is, the relationships between different local features extracted from a single video sequence frame. We now explain how 3-dimensional geometry is learnt. The key idea revolves around the descriptor transition table representation which plays the central role in our foreground/background segmentation and the estimation of the likelihood that two sequences (gallery and query) contain the same object.

Consider an image of an object overlaid with a dense grid of overlapping image patches, such as the one previously introduced in Fig~\ref{f:grid}, and within it a particular patch at the location $(x,y)$ with the corresponding descriptor word $w_0$. As the viewpoint is changed, the appearance of the patch at $(x,y)$ changes as well, eventually sufficiently so to correspond to an entirely different descriptor word. Depending on the direction of viewpoint variation, the word observed at $(x,y)$ may change from $w_0$ to any one of the words in the set of all descriptor words $\{ w_i \}$.

The above allows to to define what we term the descriptor transition table (DTT) which corresponds to a particular video sequence seen in training. The value in the descriptor transition table $T$ at row $j$ and column $k$ is the probability that the observed descriptor word $w_j$ makes the transition to the word $w_k$ for a small viewpoint change:
\begin{align}
T(j,k) = p(w_j \rightarrow w_k).
\end{align}
The probabilities of a transition table can be readily seen to capture the relationship between appearances of the same object from different views and thereby, implicitly, its geometry too. Broadly speaking, the spirit of the key idea here is similar to that of e.g.\ spatio-temporal interest points~\cite{YuanLiHuLing+2013} or 3D LBPs~\cite{TangYinSunHu2013}.

\paragraph{Learning a descriptor transition table} Following our definition of a descriptor transition table, it is tempting to consider changes of descriptor words between successive frames only in the estimation of the aforementioned probabilities $p(w_j \rightarrow w_k)$ which correspond to different table entries. We do not adopt this approach as it is inherently sensitive to the actual ordering of objects views observed in a particular sequence. In other words, different video sequences can contain exactly the same views of an object but ordered differently. Thus we argue that for the purpose of DTT estimation, training video frames should be treated as a set, rather than an ordered sequence.

\begin{figure}[thp]
  \centering
  \includegraphics[width=0.30\textwidth]{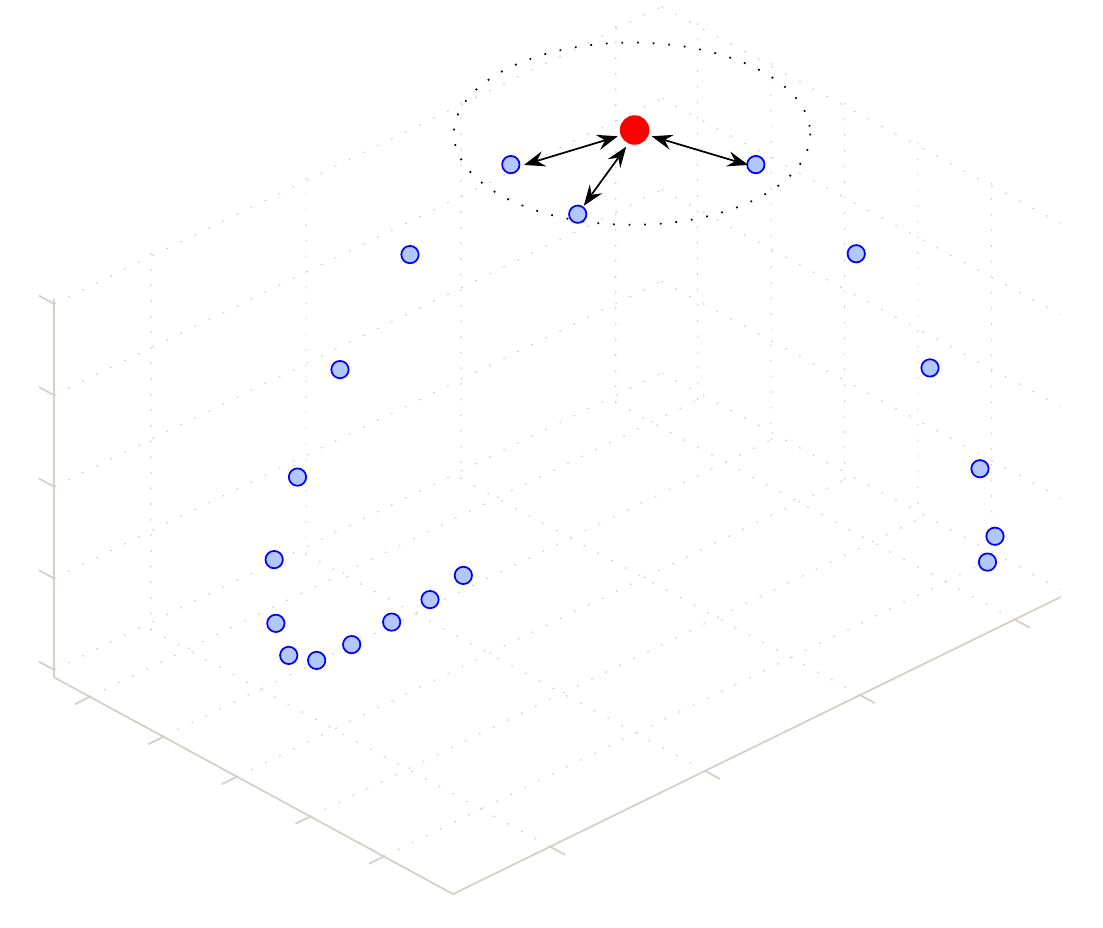}
  \caption{The proposed learning of the descriptor transition table corresponding to an object in a training video sequence does not rely on the ordering of object views in the video. Instead we consider all descriptor word transitions between all ``possibly successive'' pairs of frames (after coarse background removal) as determined by their distance in the image space.}
  \label{f:manifold}
\end{figure}

Our approach to populating a DTT from a training video sequence consists of considering all possible frame successions. We say that two frames $I_i$ and $I_j$ are ``possibly successive'' if their normalized distance in the image space is less than the threshold $t$:
\begin{align}
  \| I_i - I_j \| ~/~ \| I_i \| \leq t,
\end{align}
as illustrated in Fig~\ref{f:manifold} (in this work we used $t=0.1$). Not only does this approach accomplish the desired independence of view sequencing \cite{AranCipo2013} but it also has the advantage of using in the estimation multiple transitions per frame, rather than only a single one actually observed.

\begin{figure}[!t]
  \centering
  \subfigure[Descriptor word transitions inference]{\includegraphics[width=0.40\textwidth]{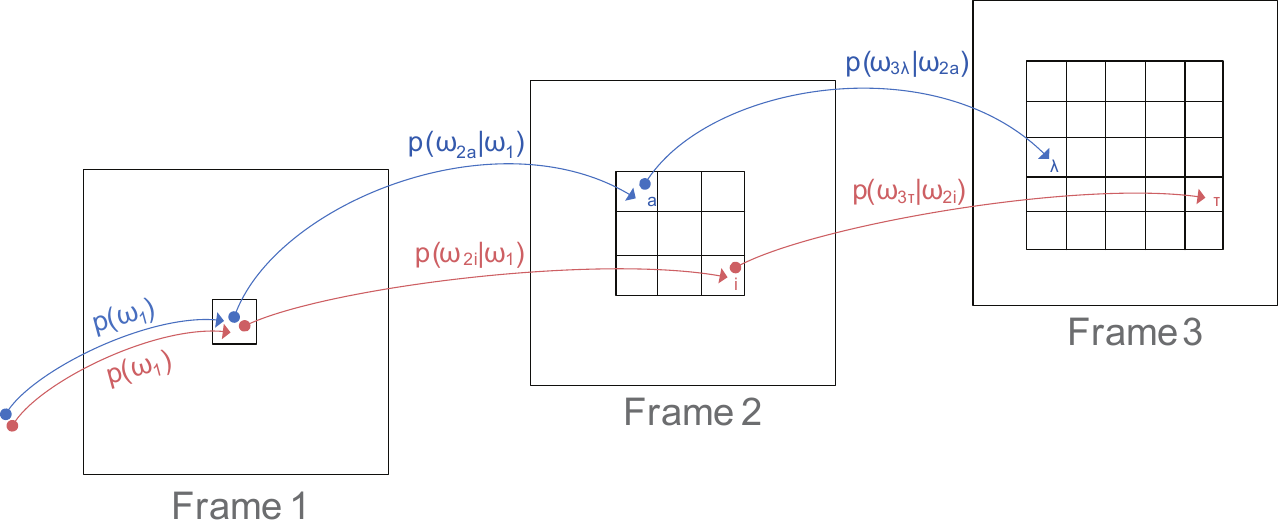}}\\[5pt]
  \subfigure[Segmentation using coherence of transition paths]{\includegraphics[width=0.40\textwidth]{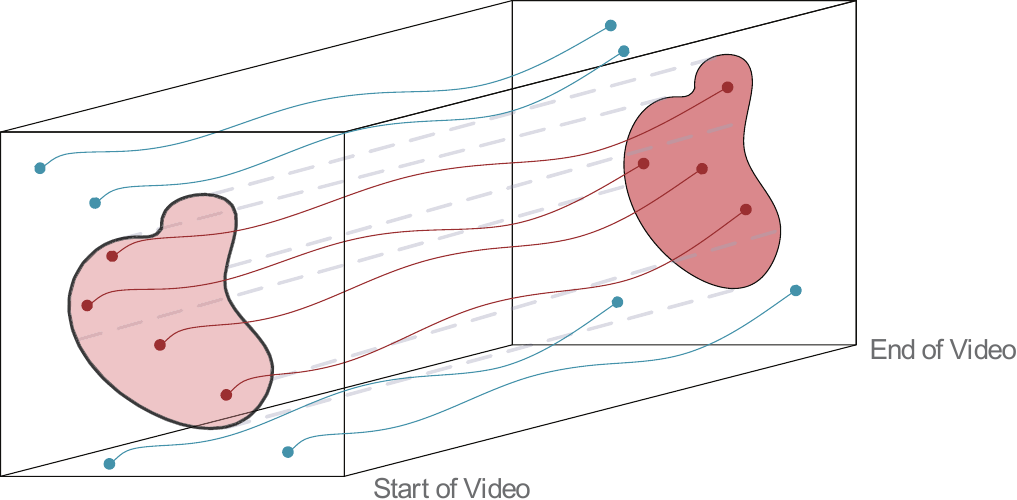}}
  \caption{Descriptor word transitions: (a) possible transitions are considered in frame-to-frame $9 \times 9$ neighbourhood (using the grid illustrated in Fig~\ref{f:grid} and defined in Sec~\ref{sss:geoFrame}), and (b) word trajectories through the time dimension are used to perform Graph Cuts based quasi-volumetric segmentation of foreground/background in a video; see Fig~\ref{f:segm}. }
  \label{f:transitions}
\end{figure}

\paragraph{Applying the DTT model} We now wish to apply the learnt object appearance model in the form of a descriptor transition table, to a novel video sequence of an unknown object. We treat each track of descriptor words through a video sequence as a first order Markov chain, where the probability of observing a word $w_i$ at ``time'' $n+1$ in the chain is governed by the learnt DTT:
\begin{align}
  p(X_{n+1} = w_i | X_n = w_j) = T(j, i).
\end{align}
The track starting in the first frame at the word $X_0 = w_{i(0)}$ is then produced by maximizing the likelihood:
\begin{align}
  p(w_{i(1)} | w_{i(0)}) ~p(w_{i(2)} | w_{i(1)}) \ldots p(w_{i(N)} | w_{i(N-1)})
  \label{e:mchain}
\end{align}
under the ``bound velocity'' constraint on patch correspondence:
\begin{align}
  \left\|\left(
    \begin{array}{c}
      x_{n+1} \\
      y_{n+1} \\
    \end{array}
  \right)
  -
  \left(
    \begin{array}{c}
      x_n \\
      y_n \\
    \end{array}
  \right) \right\|
  \leq d,
\end{align}
where $(x_n, y_n)$ is the location of the $n$-th patch in the chain and $d=\sqrt{2}$ (restricting the possible transition loci to the $3\times 3$ neighbourhood), as illustrated in Fig~\ref{f:transitions}. This maximization is readily achieved using dynamic programming and the well-known Viterbi algorithm.

\subsection{Quasi-volumetric segmentation}
At this stage from a query video we have produced a set of descriptor word transitions through the sequence. Let's call one such track of transitions $t_i$:
\begin{align}
    t_i = \left\{ (w_1^{(i)}, x_1^{(i)}, y_1^{(i)}), \ldots, (w_{N_i}^{(i)}, x_{N_i}^{(i)}, y_{N_i}^{(i)})
    \right\},
\end{align}
where $w_j^{(i)}$ is the word at $(x_j^{(i)}, y_j^{(i)})$ that transitions to $w_{j+1}^{(i)}$ at $(x_{j+1}^{(i)}, y_{j+1}^{(i)})$. By construction, meaningful tracks should weave through the object and not trough the background. Thus, we seek to choose optimally a subset of tracks $\mathcal{T}_o$ which explains the object's appearance.

\begin{figure}[!t]
  \centering
  \subfigure[Original image input (single frame from a sequence)]{~~~\fbox{\includegraphics[width=0.35\textwidth]{obj_01_01.jpg}}~~~}\\[5pt]
  \subfigure[Corresponding slice through quasi-volumetric segmentation]{~~~\fbox{\includegraphics[width=0.35\textwidth]{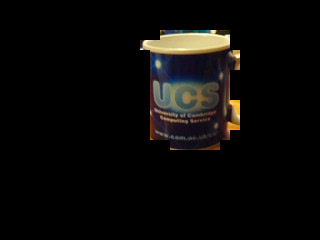}}~~~}
  \vspace{5pt}
  \caption{(a) Typical frame from a raw video sequence, and (b) the same frame with the background removed following the proposed quasi-volumetric foreground/background segmentation of the video.}
  \label{f:segm}
\end{figure}

Our approach uses the Graph Cuts algorithm~\cite{Prin2012}. Motivated by the argument laid out above and in contrast to previous methods, we apply it on the descriptor track level. Unlike in the case when Graph Cuts is used on a single image, the potential of our tracks to diverge, interlace or intercept means that the underlying graph and its structure are not inherent in the basic elements that are being discriminated. Instead we construct it as follows:
\begin{itemize}
  \item each track corresponds to a graph node
  \item the cost of assigning the label ``background'' to the track $t_j$ is the probability of the corresponding Markov chain in \eqref{e:mchain}
  \item nodes corresponding to tracks $t_i$ and $t_j$ are connected iff in any frame the distance between the patches they pass through is less than 2 pixels:
     \begin{align}
       \exists k.~(x_k^{(i)} - x_k^{(j)})^2 + (y_k^{(i)} -
      y_k^{(j)})^2 < 2^2
     \end{align}
  \item the cost of assigning different labels to tracks $t_i$ and $t_j$ is:
    \begin{align}
      e_{ij} = \sum_k \left[ (x_k^{(i)} - x_k^{(j)})^2 + (y_k^{(i)} -
      y_k^{(j)})^2 \right]^{-1}.
    \end{align}
\end{itemize}
Following the application of Graph Cuts, the tracks labelled as foreground define a 2D+time volume which allow the object to be segmented out, as illustrated in Fig~\ref{f:segm}. Finally, the likelihood of the same object being present in the two compared sequences can be obtained by computing the likelihood in \eqref{e:mchain} using only tracks segmented out as belonging to the foreground (i.e.\ the best object hypothesis for the comparison).

\section{Evaluation}
In this section we turn our attention to the evaluation of the proposed method. We begin by describing the data which was used to train and query different algorithms, continue with a summary of the existing methods we compared our approach with, and finish with a presentation of the results and a discussion.

\subsection{Data sets}\label{ss:data}
\begin{figure}[thp]
  \centering
  \includegraphics[width=0.48\textwidth]{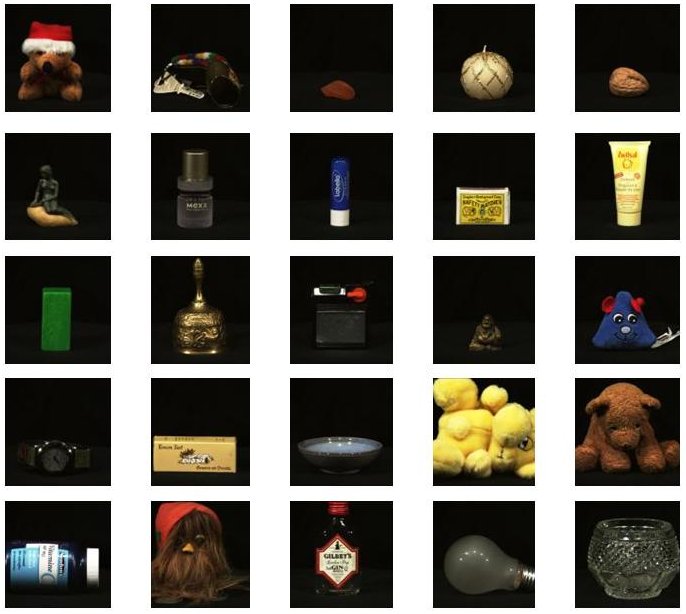}
  \caption{Examples of objects from the Amsterdam Library of Object Images. The library includes a large number of objects (1000) with varying textural and shape properties, with many objects sharing similar appearance or shape. The data set was acquired in controlled conditions (uniform viewpoint sampling, uniform background) which allowed a well-controlled evaluation protocol to be adopted as a means of gaining insight into the strengths and weaknesses of different methods. }
  \label{f:aloi}
\end{figure}

To evaluate the performance of the proposed algorithm and compare it to previously proposed approaches, we used two pertinent data sets. These are the publicly available Amsterdam Library of Object Images (ALOI)~\cite{GeusBurgSmeu2005} and a highly challenging data set of video sequences acquired using a mobile phone, collected by ourselves (the data set will be made public when anonymity is no longer required). A comprehensive description of the ALOI can be found in the original publication. In the context of the present work it suffices to summarize it by noting that the data set is very large, comprising sets of images of 1000 objects. The set of images of a particular object corresponds to 72 different viewpoints at uniformly sampled yaw values i.e.\ at successive $5^{\circ}$ rotations about the vertical axis. The ALOI contains a diverse range of objects, some of which are very much alike one another, sharing similar appearances or shapes. Examples are shown in Fig~\ref{f:aloi}. The data was acquired in controlled conditions (uniform viewpoint sampling, uniform background) which allowed us to design a well-controlled evaluation protocol as a means of gaining initial insight into the strengths and weaknesses of different evaluated methods.

\begin{figure*}[thp]
  \centering
  \subfigure[Digital clock]{\fbox{\includegraphics[width=0.225\textwidth]{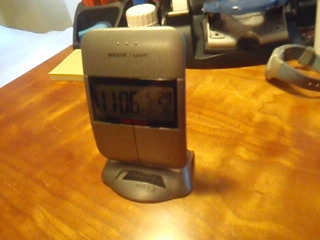}}}~~~~~
  \subfigure[Rubik's cube]{\fbox{\includegraphics[width=0.225\textwidth]{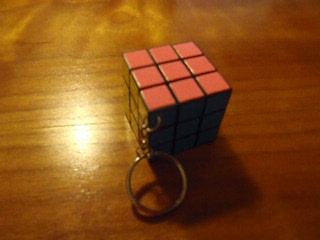}}}~~~~~
  \subfigure[Box of tissues]{\fbox{\includegraphics[width=0.225\textwidth]{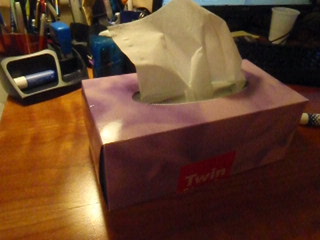}}}~~~~~
  \subfigure[Dinner plate]{\fbox{\includegraphics[width=0.225\textwidth]{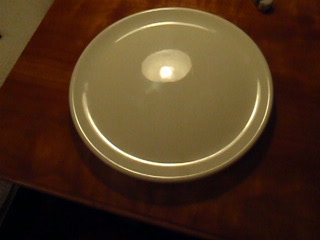}}}\\[10pt]
  \subfigure[Printer]{\fbox{\includegraphics[width=0.225\textwidth]{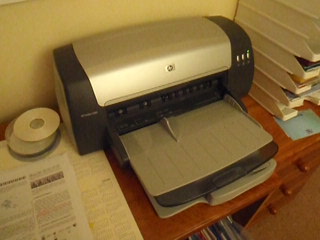}}}~~~~~
  \subfigure[Coffee mug]{\fbox{\includegraphics[width=0.225\textwidth]{obj_01_04}}}~~~~~
  \subfigure[Box of pills]{\fbox{\includegraphics[width=0.225\textwidth]{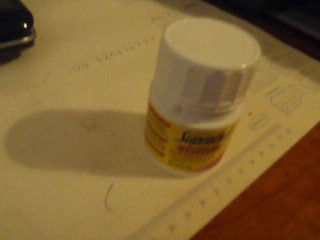}}}~~~~~
  \subfigure[Molecular model]{\fbox{\includegraphics[width=0.225\textwidth]{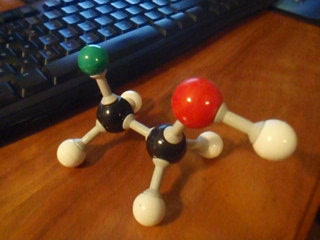}}}
  \caption{Newly collected database of object video sequences: examples. Shown are representative frames from the corresponding video sequences and a succinct description of the imaged object.}
  \label{f:ourDB}
  \vspace{10pt}
\end{figure*}

Unlike the ALOI, the second set we used for evaluation was acquired in highly uncontrolled conditions. In particular it comprises 100 video sequences acquired using a mobile phone, with 2 sequences for each of the 50 objects. Objects were imaged in a room lit by artificial lighting. The placement of an object in the two sequences was different, with major changes in background clutter, illumination, pose, scale, and camera motion. Some of the challenges were already illustrated in Fig~\ref{f:challenges}, while Fig~\ref{f:ourDB} shows some additional examples of objects in the data set. Notice that some of the objects are untextured and some ``wiry'' (e.g.\ respectively the dining plate and the molecular model in Fig~\ref{f:ourDB}). In addition to general clutter, also observe the presence of shadows as well as specular reflections. We purposefully included similar objects such as, for example, a stapler and a hole-punch.

\subsection{Baseline methods}
We consider several baseline set representations which either demonstrate state-of-the-art performance in comparable recognition tasks or which have been recently described in the literature. These are: (i) sets of SIFT local descriptors, (ii) Gaussian mixture models, (iii) linear subspaces. As usual we fit Gaussian mixtures by employing probabilistic principal component mixtures and minimizing the corresponding model+data description length; following recommendations from prior work~\cite{WolfHassMaoz2011} for the subspaces based baseline we adopt 6-dimensional subspaces.

We adopt two baseline set similarity measures, again motivated by the reports of their good performance in the existing literature. The first of these is the Kullback-Leibler divergence~\cite{KleiFrin2011} applied in the context of the Gaussian mixture based representation and estimated numerically as no analytical solution exists (we shall refer to this method as Appearance+KLD). The second similarity measure we adopted and which we applied in the context of SIFT descriptor sets and linear subspaces is the algebraic method based on the maximum correlation between pairs of vectors lying in two subspaces (we shall refer to this method as SIFT+COS), which is an extension of the \textit{maximum maximorum} (`max-max') cosine similarity between sets of exemplars  $\max_{f_1 \in S_1,f_2 \in S_2} f_1^T f_2 / \|f_1\| /\|f_2\|$ \cite{AranCipo2006e,Aran2009,Aran2016a}. In recent experiments~\cite{WolfHassMaoz2011} this method was shown to outperform a number of alternatives including by a large margin the pyramid match kernel of Grauman and Darrell~\cite{GrauDarr2005} and the locality-constrained linear coding (LLC) of Wang \textit{et al.}~\cite{WangYangYuLv+2010}. Lastly we also apply the aforementioned maximum correlation based distance on raw appearance too (Appearance+COS).

\subsection{Results and discussion}
We started our evaluation by experiments on the ALOI, designed to examine how well our algorithm copes with recognition across viewpoint changes. Generalization from a limited viewpoint range to a different, also limited viewpoint range, is a major challenge yet one that is frequently encountered in practice. We adopted the following evaluation protocol:
\begin{itemize}
  \item for all possible viewpoint angles $\alpha$ (relative to an arbitrary origin of choice), the images in the viewpoint range $(\alpha,\alpha + \Delta \phi)$ of breadth $\Delta \phi$ are used as a training sequence/set,
  \item for all possible viewpoint range shifts $\Delta \alpha$, the images in the viewpoint range $(\alpha + \Delta \alpha, \alpha + \Delta \alpha + \Delta \phi)$
        are used as the query sequence/set,
  \item the performance at the shift of $\Delta \alpha$ is quantified by the average recognition rate over all test cases.
\end{itemize}
Because all training and query sequences contain only a limited range of views, this protocol is much more challenging than when views are chosen as random subsets of the original 72 views. For clarity, all results reported in this paper were produced using $\Delta \phi = 40^{\circ}$ -- we found that the results obtained using this value are representative, qualitatively speaking, of general performance trends across different methods examined.

\begin{table*}[!t]
  \centering
  \small
  \renewcommand{\arraystretch}{1.5}
  \caption{The performance of different methods on our new data set of video sequences acquired using a mobile phone camera in the presence of major clutter, illumination, and viewpoint changes. In addition to the average rank-1 recognition rate the confidence of correct recognitions is quantified by the ratio of the similarity between the query and correct match, and the query and the second best match (Appearance+KLD and Appearance+COS algorithms recognized no object correctly so the corresponding quantity is undefined).  }
  \begin{tabular}{l|cccc}
    \Hline
    Method      & SIFT+COS & Appearance+KLD & Appearance+COS & Proposed method\\
    \hline
    Rank-1 rate & 0.06     & 0.00           & 0.00           & 1.00\\
    Separation  & 1.22     & N~/~A          & N~/~A          & 1.92\\
    \Hline
  \end{tabular}
  \label{t:resOurDB}
\end{table*}

\begin{figure}[!bht]
  \centering
  \includegraphics[width=0.48\textwidth]{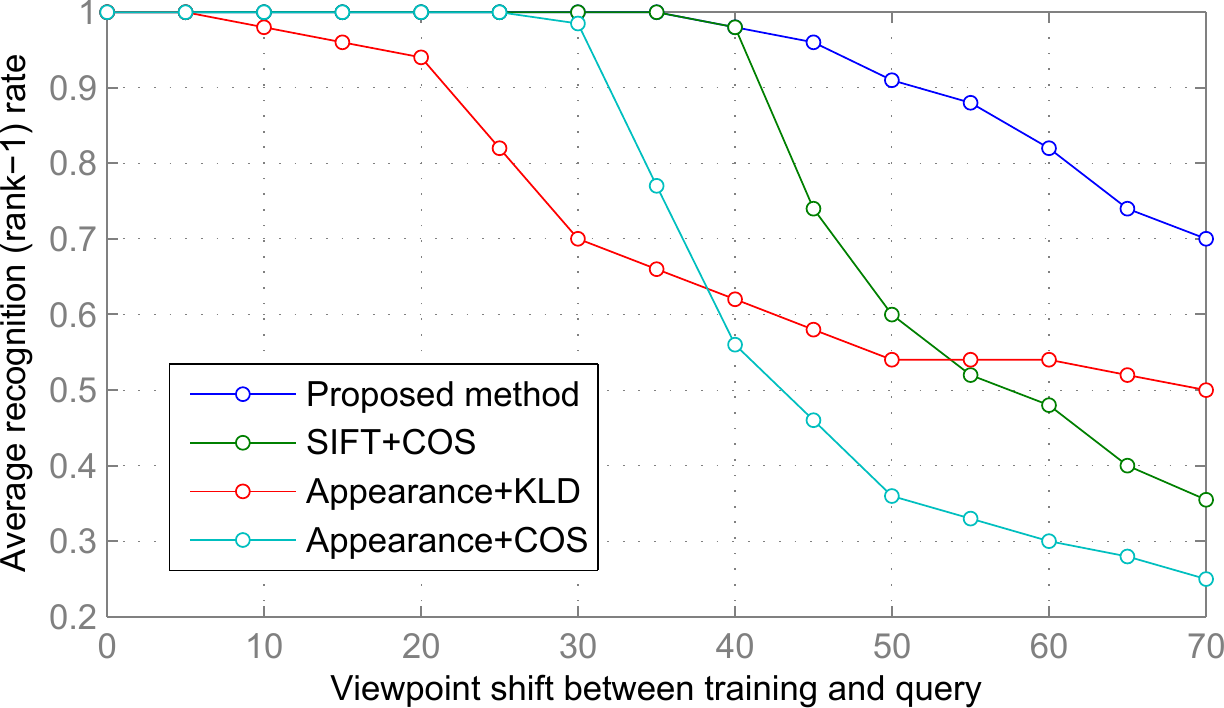}
  \caption{The average rank-1 recognition rate achieved using different methods across viewpoint changes, using image sets constructed from the Amsterdam Library of Object Images. As expected the performance of all methods deteriorates with the increase in the viewpoint difference between training and query sequences. However the proposed method demonstrates far superior behaviour than all other methods.}
  \label{f:resALOI}
  \vspace{10pt}
\end{figure}

The key results are summarized by the plot in Fig~\ref{f:resALOI} which shows the variation in the rank-1 recognition rate achieved using different methods as a function of the viewpoint change between the training and query sequences. As expected both from theory and previous reports in the literature, the performance of all methods deteriorates with an increase in the viewpoint difference between training and query sequences. The most rapid deterioration is observed for the KLD based method which highlights the inherent inability of probability density based methods to generalize -- if the training set does not contain representative variability, recognition performance for arbitrary novel input is likely to be poor. For viewpoint changes of moderate extent ($\Delta \alpha <40^\circ$) generalization is improved with subspace modelling and the use of a more invariant correlation based similarity measure (also consistent with the previous findings in the literature~\cite{WolfHassMaoz2011}), as witnessed by markedly better performance of the Appearance+COS algorithm. Interestingly this initial improvement is not maintained for large viewpoint changes of over $40^\circ$. Considering the nonlinear distribution of object appearance within the corresponding image space, deterioration for linear subspace based approaches is certainly expected, yet it is unclear why it would be any greater than for the density based KLD algorithm. The use of the SIFT descriptor, which itself has been shown to show good resilience to both illumination and viewpoint, confers further benefit, with the corresponding SIFT+COS algorithm exhibiting even slower deterioration across a wide range of viewpoint changes ($\Delta \alpha <55^\circ$, and most significantly so for $\Delta \alpha <40^\circ$). However this method too is outperformed by the simple Appearance+KLD approach for changes of over $55^\circ$. Lastly, the proposed method is readily seen to exhibit vastly superior performance in comparison with all of the other methods and across the entire range of viewpoint changes. Even for the extreme change of $70^\circ$ it attains over 70\% correct recognition rate. In comparison, the recognition rate of the Appearance+KLD approach drops to the same level already for a $30^\circ$ viewpoint differential, for Appearance+COS for $37^\circ$, and SIFT+COS for $46^\circ$.

Following the highly promising findings on the ALOI in terms of the superiority of the proposed method, we next sought to evaluate how the algorithms perform on truly realistic video sequences and the newly introduced data set described in Sec~\ref{ss:data}. We used one of the image sequences of an object for training, and the other one (recall, in a different context, with changes in background clutter, viewpoint, camera motion, and illumination) as query. As before we initially examined the rank-1 recognition rate of different algorithms, that is, the rate at which the correct gallery sequence was found to be the best match to the query. The results are summarized in Table~\ref{t:resOurDB} (first data row). It can be immediately seen that the superiority of our algorithm over the evaluated alternatives is demonstrated best in highly challenging conditions such as those present in this data set. Our algorithm correctly identified the query object in all cases, thereby achieving perfect recognition performance. In contrast, the two appearance based approaches (Appearance+KLD and Appearance+COS) recognized none of the objects correctly, with the SIFT based algorithm coping with the challenges somewhat better but still poorly in comparison with the proposed method.

Lastly, to assess the confidence of the successful recognitions, when a successful recognition is observed we examined the ratio of the likelihoods corresponding to the top (i.e.\ the correct) match and the second best match (which is by implication incorrect). The results can be found in Table~\ref{t:resOurDB} (second data row). Since Appearance+KLD and Appearance+COS methods recognized no object correctly, no measurement could be taken. Comparing the results of the proposed method and that of SIFT+COS approach, we can see that not only did our algorithm exhibit vastly superior performance in terms of rank-1 recognition but also that its correct decisions were made with much greater confidence (over 50\% greater class separation).

\section{Conclusions}
We described a novel method for object recognition that uses video sequences both as training and query input. The main novelty lies in the framework used to employ
discretized local features to describe a video sequence, as well as the manner in which such features are selected in the matching process. One of the key ideas is that of the descriptor transition table which implicitly captures the 3D geometry of an object by considering the transition of a local feature from one descriptor word to another as the camera viewpoint changes. The proposed method was demonstrated as effective an in empirical evaluation on the publicly available Amsterdam Library of Object Images and a new highly challenging data set of video sequences acquired using a mobile phone.

\balance

\bibliographystyle{ieeetran}
\bibliography{../../../my_bibliography}

\end{document}